\documentclass[runningheads]{llncs}
\usepackage{graphicx}
\usepackage{amsmath,amssymb} 
\usepackage{todonotes}
\usepackage{hyperref}
\usepackage{mathtools}

\newcommand{\arguments}{\ensuremath{\mathcal{A}}}
\newcommand{\attacker}{\ensuremath{\mathrm{Att}}}
\newcommand{\supporter}{\ensuremath{\mathrm{Sup}}}
\newcommand{\baseScore}{\ensuremath{\beta}}
\newcommand{\strength}{\ensuremath{\sigma}}

\newcommand{\cinput}{\ensuremath{\mathbf{x}}}
\newcommand{\coutput}{\ensuremath{y}}
\newcommand{\argsin}{\ensuremath{\mathcal{A}_{\textrm{in}}}}
\newcommand{\argsout}{\ensuremath{\mathcal{A}_{\textrm{out}}}}
\newcommand{\argsl}[1]{\ensuremath{\mathcal{A}_{\langle#1\rangle}}}

\begin{document}
\title{Learning Gradual Argumentation Frameworks using Genetic
Algorithms}
%
%
\author{Jonathan Spieler\orcidID{0000-0002-9767-4989} \and
Nico Potyka\orcidID{0000-0003-1749-5233} \and
Steffen Staab\orcidID{0000-0002-0780-4154}  }
\authorrunning{J. Spieler, N. Potyka, S. Staab}
%
\institute{University of Stuttgart, Germany,
\email{st169996@stud.uni-stuttgart.de}
}
\maketitle              
\begin{abstract}
Gradual argumentation frameworks represent arguments and their relationships
in a weighted graph. Their graphical structure and intuitive semantics makes
them a potentially interesting tool for interpretable machine learning. It has been noted recently that their mechanics are closely
related to neural networks, which allows learning their weights from data by
standard deep learning frameworks. As a first proof of concept, we propose a genetic algorithm to
simultaneously learn the structure of argumentative classification models. 
To obtain a well interpretable model, 
the fitness function balances sparseness and accuracy of the classifier.
We discuss our algorithm and present first experimental results on standard benchmarks from the UCI machine learning repository. 
Our prototype learns argumentative classification models that are
comparable to decision trees in terms of learning performance and
interpretability. 

\end{abstract}
\section{Introduction}

The basic idea of abstract argumentation frameworks is to represent arguments
and their relationships in a graph.
An abstract argument can be any entity whose state can be determined solely 
by its relationships to other arguments \cite{dung1995acceptability}.
Our focus here is on bipolar frameworks that consider attack and support relationships between arguments \cite{boella2010support}.
In the context of a classification problem, where we want to assign a class
label to an instance based on its features, arguments can correspond to
class labels, feature values that fall in a particular value range and meta-arguments that are affected by more basic arguments. For example,
in a medical diagnosis example, there may be some symptoms that support and
other symptoms that attack a diagnosis. 

The idea of argumentative classification has been proposed before
in \cite{thimm2017towards}. The authors proposed applying rule mining
algorithms first to extract interesting arguments from data and to 
feed those in a structured argumentation engine. While the idea is
conceptually very interesting, the problem is that the rule learning step
and the classification step cannot be connected easily, so that often too
many meaningless rules are learnt. 
Therefore, quantitative argumentation frameworks have been proposed as
an alternative because they directly allow regarding features as inputs
and training the classification model in an end-to-end fashion
from data \cite{Potyka20}.
Gradual abstract argumentation frameworks 
\cite{amgoud2017acceptability,amgoud2008bipolarity,baroni2015automatic,potyka2018Kr,rago2016discontinuity} are particularly interesting for this purpose
because their mechanics are very close to neural networks, which allows
learning their weights by standard deep learning frameworks \cite{Potyka21}.
However, in this context, usually fully connected graphs are considered.
Our focus here is, in particular, on learning sparse graphs that make the
model better interpretable. We propose a genetic algorithm that 
constructs argumentative classifiers in an end-to-end fashion from data.

We explain the necessary basics of gradual argumentation frameworks in
the next section. We describe a genetic algorithm for learning
gradual argumentation frameworks from data in Section \ref{sec_gen_alg}
and present some experimental results in Section \ref{sec_experiments}.

\section{Background}
\label{sec_background}

\begin{figure}[tb]
	\centering
		\includegraphics[width=0.99\textwidth]{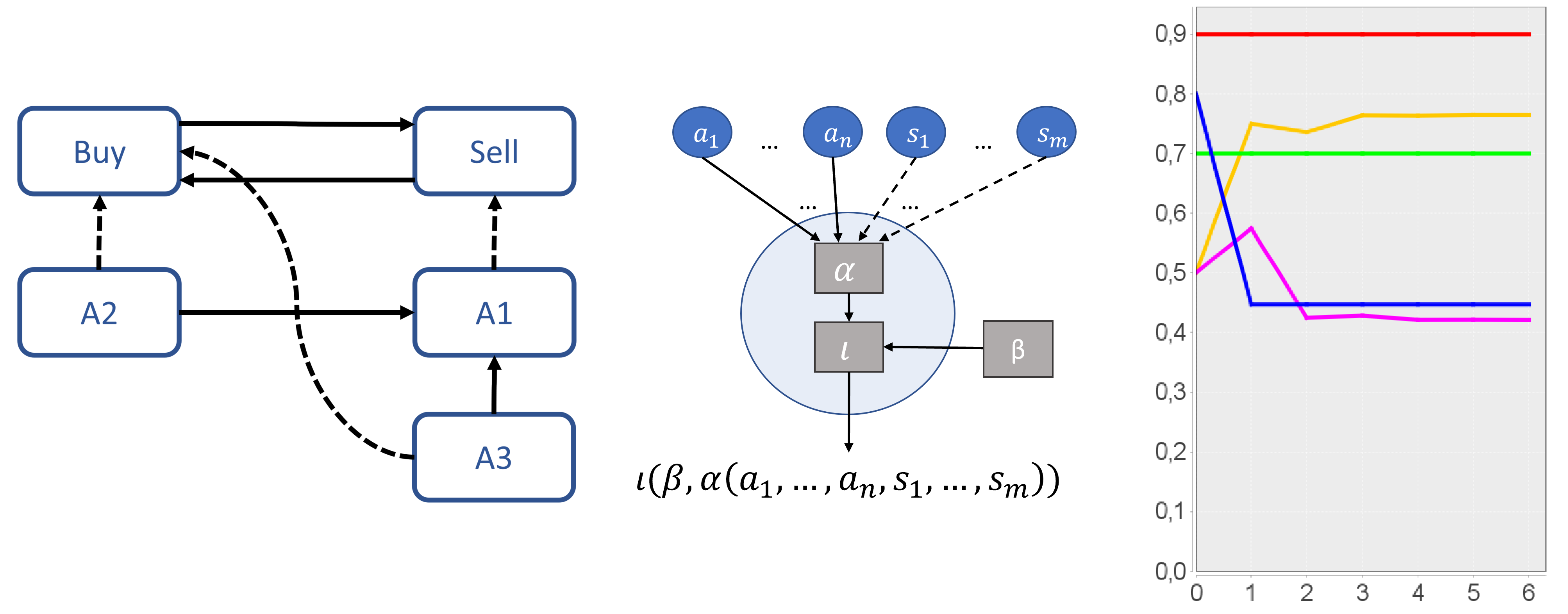}
	\caption{Example of a GAF (left), illustration of update mechanics under modular semantics (middle) and evolution of strength values under MLP-based semantics (right).}
	\label{fig:fig_GAF}
\end{figure}

We consider gradual argumentation frameworks (GAFs for short) consisting of 
an argumentation
graph whose nodes correspond to abstract arguments and edges to attack and support relationships between the arguments. For our purposes, an argument
is just an abstract entity that can be accepted or rejected to a certain degree
based on the state of its attackers and supporters. Every argument is
associated with a base score that can be seen as its apriori strength when
ignoring its attackers and supporters. We consider edge-weighted
GAFs similar to \cite{mossakowski2018modular}.
\begin{definition}[Gradual Argumentation Framework (GAF)]
A GAF is a tuple
$(\arguments, E, \baseScore, w)$ that consists of
\begin{itemize}
    \item a set of arguments $\arguments$ and a set of edges $E \subseteq \arguments \times \arguments$
between the arguments,
    \item a function $\baseScore: \arguments \rightarrow [0,1]$
that assigns a \emph{base score}
to every argument and
    \item a function $w: E \rightarrow [-1,1]$
     that assigns a weight to every edge.
\end{itemize}
Edges with negative weights are called \emph{attack} and edges with positive 
weights are called \emph{support} edges and denoted by $\attacker$ and 
$\supporter$, respectively.
\end{definition}
Figure \ref{fig:fig_GAF} shows on the left the graphical structure of a GAF 
that formalizes a simple decision 
problem. We assume that we want to decide if we should buy or sell
stocks of a company and that we consider three arguments put forward
by different experts. Attack relationships are denoted by solid and
support relationships by dashed edges.
We can define the semantics of GAFs by interpretations that assign a strength
value to every argument. 
\begin{definition}[GAF interpretation]
\label{def_QBAF_interpretation}
Let $G = (\arguments, E, \baseScore, w)$ be a GAF.
An interpretation of $G$ is a function $\strength: \arguments \rightarrow [0,1] \cup \{\bot\}$
and $\strength(a)$ is called the strength of $a$ for all $a \in \arguments$.
If $\strength(a) = \bot$ for some $a \in \arguments$, $\strength$ is called \emph{partial}. Otherwise,
it is called \emph{fully defined}.
\end{definition}
Interpretations are often defined based on an iterative procedure that
initializes the strength values of arguments with their base scores and
repeatedly updates the values based on the strength of their attackers and supporters. Interpretations are partial when the procedure fails to converge. As shown in \cite{mossakowski2018modular}, this can indeed happen in cyclic graphs.
However, we will only be concerned with acyclic graphs here and all interpretations will be fully defined.

The strength update can often be divided into two main steps \cite{mossakowski2018modular}: first, an \emph{aggregation function} $\alpha$ 
aggregates the strength values of attackers and supporters.
Then, an \emph{influence function} $\iota$ adapts the
base score based on the aggregate as illustrated in Figure \ref{fig:fig_GAF}
in the middle. 
Examples of aggregation functions include product \cite{baroni2015automatic,rago2016discontinuity},
addition \cite{amgoud2017evaluation,potyka2018Kr} and maximum \cite{mossakowski2018modular} and the influence function
is defined accordingly to guarantee that strength values fall
in the desired range. 
Semantical guarantees of these approaches have been discussed in \cite{amgoud2017evaluation,baroni2018many,potyka2018Kr}.
 
We will only look at the MLP-based semantics from \cite{Potyka21} here.
Under this semantics, layered acyclic GAFs actually correspond to 
multilayer perceptrons (MLPs), a popular class of neural networks.
The strength values of arguments are computed by the following iterative
procedure:
For every argument $a \in \arguments$, we let $s_a^{(0)} := \baseScore(a)$
be the initial strength value. The strength values are then updated by doing the following two steps repeatedly for all
$a \in \arguments$:
\begin{description}
\item[Aggregation:] We let $\alpha_a^{(i+1)} := \sum_{(b,a) \in E} w(b,a) \cdot s_b^{(i)}$.
\item[Influence:] We let $s_a^{(i+1)} := \varphi_l\big(\ln(\frac{\baseScore(a)}{1- \baseScore(a)}) + \alpha_a^{(i+1)} \big)$,
where $\varphi_l(z) = \frac{1}{1 + \exp(-z)}$ is the logistic function.
\end{description} 
Figure \ref{fig:fig_GAF} shows on the left how the strength values of
our running example evolve under MLP-based semantics. 
The weight of all attack (support) edges was set to $-1$ (1) and the base 
scores correspond to the strength values (y-axis) at iteration 0 (x-axis).

\begin{figure}[tb]
	\centering
		\includegraphics[width=0.65\textwidth]{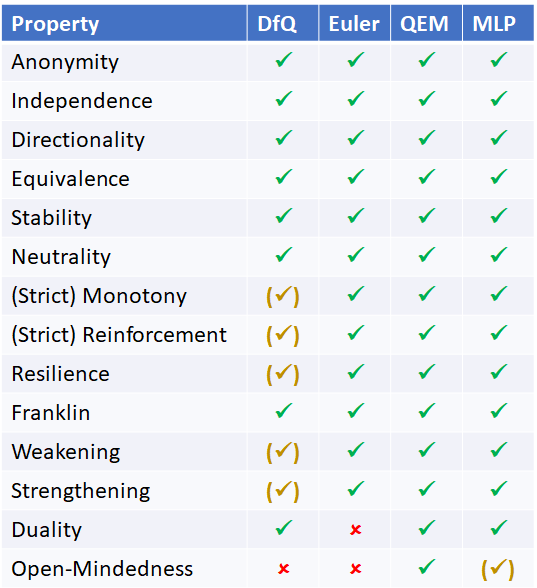}
	\caption{Desirable properties for gradual argumentation semantics. 
	$\checkmark$ indicates full satisfaction,
	$(\checkmark)$ satisfaction if base scores $0$ and $1$
	are excluded and 
	$\textit{x}$ indicates that the 
	property can be violated even when 
	excluding base scores $0$ or $1$.
	\label{fig:properties}
    }
\end{figure}

As shown in \cite{Potyka21}, the MLP-based semantics satisfies almost all
semantical properties from the literature perfectly. 
A comparison to the earlier semantics 
DF-QuAD \cite{rago2016discontinuity}, Euler-based semantics
\cite{amgoud2017evaluation}
and the quadratic energy model \cite{potyka2018Kr} are shown in Table \ref{fig:properties}.
We refer to \cite{Potyka21} for more details.
Only the property \emph{Open-Mindedness} can be violated
when arguments have base scores $0$ or $1$. 
In this case, the strength values remain necessarily at 
$0$ or $1$ under MLP-based semantics.
Roughly speaking, \emph{Open-Mindedness} demands that
continuously adding attackers/supporters to an argument will eventually bring the strength to $0$/$1$ \cite{Potyka19om}.
For readers familiar with some, but not all of the properties,
let us note that \emph{Open-Mindedness} is similar to some "monotonicity"-properties that demand that an attacker/supporter
must decrease/increase the strength of an argument.
However, these properties are not sufficient to guarantee that
the strength converges to $0$/$1$ in the limit. For example,
the Euler-based semantics from \cite{amgoud2017evaluation} satisfies these monotonicity 
properties in most cases, but the lower limit is not $0$,
but $\beta(a)^2$ (base score squared) \cite{Potyka19om}.
Let us also note that the fact that \emph{Open-Mindedness}
is not satisfied for base scores $0$ or $1$ cannot cause
any problems in out learning setting.
This is because they correspond
to infinite weights of the corresponding MLP that cannot be taken in practice. We refer to \cite{Potyka21} for more details. The different semantics shown in Table \ref{fig:properties} can be compared experimentally
with the Java library Attractor\footnote{\url{https://sourceforge.net/projects/attractorproject/}} \cite{Potyka2018_tut}.

\section{A Genetic Algorithm for Learning GAFs}
\label{sec_gen_alg}

Our goal is to learn interpretable \emph{GAFs} that can solve classification problems.
The goal of classification is to map inputs $\cinput$ to outputs $\coutput$. A typical example is classifying
a customer as credit-worthy or not credit-worthy (output) 
based on personal data like age and income (input).
We think of the inputs as feature tuples $\cinput = (x_1, \dots, x_k)$, where the i-th value is taken from some domain $D_i$.
The output $\coutput$ is taken from a finite set of class labels $L$.
A classification problem $P = ((D_1, \dots, D_k), L, E)$ consists of the domains, the class labels and a set of training examples $E = \{(\cinput_i, \coutput_i) \mid 1 \leq i \leq N\}$. 

A numerical classifier is a function $c: \big(\bigtimes_{i=1}^k D_i\big) \times L \rightarrow \mathbb{R}$
that assigns to every pair $(\cinput, \coutput)$ a numerical value. An important special case is a probabilistic classifier
$p: \big(\bigtimes_{i=1}^k D_i\big) \times L \rightarrow [0,1]$ where $\sum_{j=1}^{|L|} p(\cinput, \coutput_i) = 1$.
Then $p(\cinput, \coutput) \in [0,1]$ can be understood as the confidence of the classifier that an example with features $\cinput$ belongs to 
the class $\coutput$. Note that every numerical classifier $c$ can be turned into a probabilistic classifier $p_c$ by normalizing the label
outputs by a softmax function. That is, $p_c(\cinput, \coutput) = \frac{\exp(c(\cinput, \coutput_i))}{\sum_{j=1}^{|L|}\exp(c(\cinput, \coutput_j))}$. 

\begin{figure}[tb]
	\centering
		\includegraphics[width=0.99\textwidth]{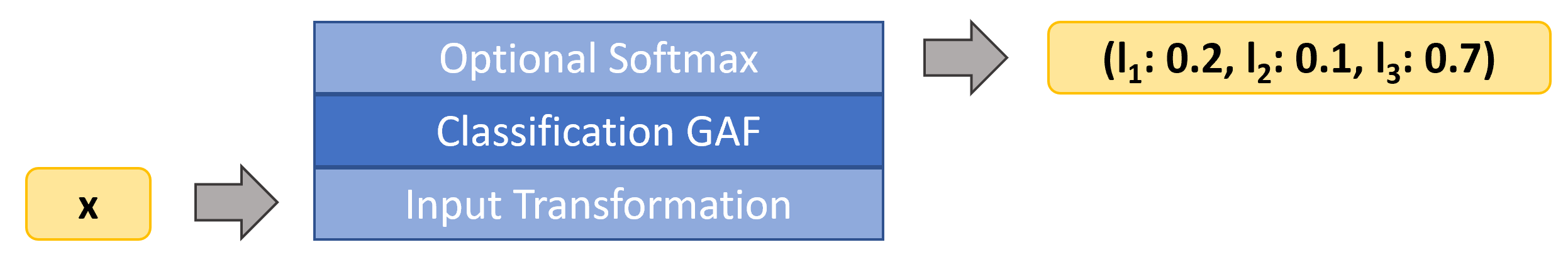}
	\caption{High-level architecture of classification GAFs adapted from \cite{Potyka20}.
	\label{fig:class_gaf}
    }
\end{figure}

Following \cite{Potyka20}, we are interested in learning
argumentative classifiers with a high-level structure as
illustrated in Figure \ref{fig:class_gaf}.
We focus on layered acyclic GAFs, where the first layer is composed of \emph{input arguments} that correspond to
the input features and the last layer
is composed of \emph{output arguments} that correspond to the classes that we want to predict. 
Intermediate layers contain \emph{meta-arguments} that
can combine primitive arguments in earlier layers to
increasingly complex arguments inspired by the idea of
deep learning \cite{lecun2015deep}. 
Formally, a classification GAF has the following structure.
\begin{definition}[Classification GAF \cite{Potyka20}]
A \emph{Classification GAF with $k$ layers} for a classification problem $P = ((D_1, \dots, D_k), L, E)$
is a GAF $(\arguments, E, \baseScore, w)$ such that $\arguments = \bigcup_{i=0}^{k+1} \argsl{i}$ consists
of the input arguments $\argsl{0} = \argsin$ and output arguments $\argsl{k+1} = \argsout$ for $P$ 
and additional layers of arguments $\argsl{i}$ such that $\argsl{i} \cap \argsl{j} = \emptyset$ for $i \neq j$. 
Furthermore, $E \subseteq \bigcup_{i=0}^{k} \bigcup_{j=i+1}^{k+1} \big( \argsl{i} \times \argsl{j} \big)$,
that is,
edges can only be directed towards deeper layers.  
\end{definition}
In general, an input transformation may be necessary to
bring the inputs in an appropriate form.
Binary features can be seen as input arguments immediately.
We binarize non-binary features to turn them into abstract arguments that can be accepted or rejected.
While this introduces a learning bias,
it can also improve interpretability. 
For example, a numerical age
feature could be replaced by binary features that correspond to interesting 
age periods like teenager or middle-aged. 

Let us note that a classification GAF without any intermediate layer just corresponds to a logistic regression model with binarized features. 
Additional meta-arguments in intermediate layers allow, in particular, 
capturing non-linear dependencies. 
General \emph{classification GAFs} under MLP-based semantics can be seen as
MLPs with binary features. However, while MLPs are usually dense
(every node is connected to every node in the next layer),
our focus here is on learning a sparse and well
interpretable network structure. 
Given the graphical structure of a \emph{classification GAF}, we can
train the weights using the usual backpropagation procedure for neural 
networks that is implemented in popular libraries like Tensorflow or Pytorch.
In order to learn the graphical structure, we propose a genetic algorithm.

Genetic algorithms are nature-inspired search algorithms \cite{sastry2005genetic}. 
During the search process, they maintain a set of candidate solutions
in the search space.
The candidate solutions are called \emph{chromosomes} and the set of candidates  the \emph{population}. 
In our application, chromosomes encode argumentation 
graphs. A simple genetic algorithm template is shown in Figure \ref{fig:genetic_algorithm}.
Starting from an initial (often random) population,
the population is evolved. To begin with, a subset of the 
population (mating pool) is selected for reproduction by means of a 
\emph{fitness function} and a \emph{selection function}.
Offspring is created by applying a \emph{recombination function} that creates
a new chromosome by combining features of the parent chromosomes.
The reproduction step is followed by a \emph{mutation step} that is supposed
to move to interesting new regions of the search space. This is often
achieved by random perturbations of the chromosomes.
Finally, a \emph{replacement function} replaces part of the current population
with promising offspring. The algorithm continues until a termination criterion is met.
\begin{figure}[t]
    \centering
    \begin{align*}
		&\mathrm{population} \leftarrow \mathrm{initialize(N)} \\[0.1cm]
		&\textbf{do} \\[0.cm]
			&\hspace{0.3cm} \mathrm{mating\_pool} \leftarrow \mathrm{select}(\mathrm{population}) \\[0.1cm]
			&\hspace{0.3cm} \mathrm{offspring} \leftarrow \mathrm{recombine}(\mathrm{mating\_pool}) \\[0.1cm]
			&\hspace{0.3cm} \mathrm{offspring} \leftarrow \mathrm{mutate}(\mathrm{offspring}) \\[0.1cm]
			&\hspace{0.3cm} \mathrm{population} \leftarrow \mathrm{replace}(\mathrm{population},\mathrm{offspring}) \\[0.1cm]
		&\textbf{until} \ \textit{termination condition reached} \\[0.1cm]
		&\textbf{return} \ \mathrm{best}(\mathrm{population})
	\end{align*}
    \caption{A basic genetic algorithm template.}
    \label{fig:genetic_algorithm}
\end{figure}

In the following sections, we will describe the building blocks of our genetic algorithm in more detail. 

\subsection{Chromosome Representation}

Conceptually, chromosomes represent the graphical structure of a GAF by an 
adjacency matrix. Since we are interested in learning directed layered graphs,
we consider one row for each pair of subsequent layers that indicates
which nodes are connected. More precisely, the connections between layer $i$
and layer $i+1$ are stored in a row vector $r^i$ of length $s_i \cdot s_{i+1}$,
where $s_{i}$ ($s_{i+1}$) is the number of nodes in layer $i$ ($i+1$).
Chromosomes are formed by concatenating the rows.
Without any meta-layer, our classification 
GAF would be equivalent to a logistic regression model that can only
learn linear relationships between the features. 
Universal function approximation theorems for MLPs imply that a single 
hidden layer allows to approximate very general classes of functions if
there is a sufficient number of nodes \cite{lecun2015deep}. 
Deeper graph structures often work better in practice, but they can also 
become more difficult to understand. To get a well interpretable model and
to keep the search space small, we restrict to structures composed of 
the input layer followed by one meta-layer and one output layer here. The strength values of output arguments are
normalized by a softmax function such that their values sum up
to $1$.

\subsection{Fitness Function}
To evaluate a chromosome, the graphical structure is first interpreted as a MLP
and the parameters (base scores and edge weights) are learnt 
from the dataset. 
The MLP can be transformed into a GAF under MLP-based semantics later 
by interpreting edges with positive weights as
supports and edges with negative weights as attacks \cite{Potyka21}.
Our fitness function evaluates the GAF based on
classification accuracy (percentage of correctly classified instances) 
and a regularization term that encourages a sparse graphical structure.
The regularization term is $0$ if the graph is fully connected
and increases as the graph becomes sparser. The maximum $1$ is obtained
only for a graph without any edges. Intuitively, the regularization term 
measures the sparsity of the graph.
Our fitness function is a convex combination of accuracy and the regularization
term. 
\begin{equation}\label{eq:fitness}
    f_i = (1 - \lambda) \cdot \text{Accuracy}(\mathcal{D}\textsubscript{train}) + \lambda \cdot \frac{N_{poss., conn.} - N_{conn.}}{N_{poss., conn.}} 
\end{equation}
$N_{poss., conn.}$ is the number of possible connections based on the graph structure and $N_{conn.}$ is the actual number of connections.
The hyperparameter $\lambda \in [0,1]$ indicates the relative importance of an individual's sparsity in relation to its accuracy on the training data.
Since both accuracy and the regularization
term yield values between $0$ and $1$, our fitness function returns values between $0$ and $1$ for every choice of $\lambda$.

\subsection{Selection for Mating Pool}
There are different selection methods for genetic algorithms, e.g. roulette-wheel sampling, rank selection and $q$-tournament selection \cite{Baeck_1996,Goldberg_1991}. 
We use $q$-tournament selection since it often results in better diversity within the population compared to fitness-proportional selection and helps to avoid \textit{premature convergence} \cite{Mitchell1998genetic}.
q-tournament selection picks $q$ individuals at random from the population. 
The individual with the highest fitness among them is then selected as a parent. This procedure is repeated until the new population is complete. 
We used $q = 3$. 


\subsection{Recombination}
The main idea of the recombination operator is to improve the average quality of a population by forming new individuals from beneficial combinations of their parents \cite{Baeck_1996,Goldberg_1989}. 
It is often worthwhile to design problem-specific recombination operators,
but as a proof of concept, we just apply the $k$-point crossover operator here.
Roughly speaking, it picks two individuals at random from the selected parents. Then, $k$ random crossover points are selected in the chromosomes and the segments between these points are exchanged between the parents to form new offspring \cite{Mitchell1998genetic}. We used $k=2$ as it 
is a good compromise between offspring that can be quite similar to their parents when using single-point crossover and offspring that can be quite different from their parents for uniform crossover.


\subsection{Mutation}

Intuitively, mutation allows avoiding local optima by jumping to new states 
and is important to regain variation in case of premature convergence \cite{Baeck_1996,Goldberg_1989}. There are many different mutation operators depending on the problem and its representation \cite{Abdoun_2012}. Since our chromosomes are binary encoded, we use flip mutation. This means, that a $0$ is flipped to a $1$ and vice versa \cite{Abdoun_2012,Goldberg_1989}. Using flip mutation with a small mutation probability can be a good compromise between avoiding premature convergence as well as too many connections which lead to a decrease in interpretability.

\subsection{Replacement of Old Population}

Since for most genetic algorithms the population size is constant, a strategy for so-called \textit{survivor selection} or \textit{population replacement} is needed to decide which individuals are used for the next generation \cite{Mitchell1998genetic}. Often, these strategies use the fitness values or the age. We decide to use \textit{elitist selection} that directly passes a certain percentage of the best individuals to the next generation without changing them. This guarantees that the best fitness does not decrease over the generations \cite{Baluja_1995,Whitley_2018}. To avoid reducing diversity within the population too much, the elitist selection percentage is chosen rather low. The remaining and overwhelming part of the new population is created by the offspring.

\subsection{Termination Condition}
The newly created population forms the next \textit{generation} of chromosomes until the termination criterion is reached \cite{Mitchell1998genetic}. The algorithm is stopped either if there is no further improvement (within some \textit{tolerance)} in the fitness of the population for a pre-defined number of generations (\textit{patience}) or if an absolute number of generations is reached \cite{Michalewicz_1994}. Finally, the algorithm returns the best chromosome in the population.

\section{Experiments}
\label{sec_experiments}

\subsection{Datasets}
The performance of argumentation classifiers is evaluated on three different data sets from the UCI Machine Learning Repository, namely the Iris\footnote{\url{https://archive.ics.uci.edu/ml/datasets/iris}} (4 numerical features, 150 instances), 
the Adult Income\footnote{\url{https://archive.ics.uci.edu/ml/datasets/adult}} (14 mixed features, 48,842 instances) and 
the Mushroom data set\footnote{\url{https://archive.ics.uci.edu/ml/datasets/mushroom}} (22 categorical features, 8,124 instances).

\subsection{Baselines}
For comparison, logistic regression and decision tree classifiers are used. Logistic regression is considered to ensure that there is any benefit at all in terms of performance to use GAFs, which can be interpreted as multilayer perceptrons. Decision trees are relatively easy to interpret if they are not too deep and do not have too many nodes. Therefore, two different decision trees are considered. One, whose best parameter set was determined by a grid search and a second, whose depth was limited to obtain a comparable performance to an argumentation classifier. 

\subsection{Hyperparameter Settings and Implementation}

When computing the fitness function, the parameters are learnt by Backpropagation \cite{Rumelhart_1986}. We use cross-entropy as the loss function and Adam \cite{Kingma_2017} as the optimizer. Since the number of training epochs required depends on the structure of the classifier, \emph{Early Stopping} \cite{Prechelt_1998} is used to avoid overfitting.

 For the genetic algorithm, especially the population size and the number of generations are two important parameters. For our algorithm, the number of initial (random) connections $N_{conn.}$ between the layers and the regularization parameter $\lambda$ are important as well. Due to the large search space, restricting the number of initial connections helps to start the search within a reasonable (in terms of interpretability) range. 
 Table \ref{tab:hyperparameter_settings} shows the hyperparameter settings for the different data sets. 
\begin{table}
\centering
\small
\caption{Hyperparameter settings for argumentation classifiers on different data sets}
\label{tab:hyperparameter_settings}
\resizebox{\textwidth}{!}{%
\begin{tabular}{lrrr}
\hline
\textbf{Hyperparameter} & \textbf{Iris data set} & \textbf{Adult Income data set} & \textbf{Mushroom data set} \\ \hline
Population size & 20  & 100  & 100  \\
Generations & 20  & 20  & 20 \\
Crossover rate & 0.9  & 0.9 & 0.9 \\
Mutation rate & $1\mathrm{e}{-3}$  & $1\mathrm{e}{-3}$ & $1\mathrm{e}{-3}$ \\
Elitist percentage & 0.1  & 0.1  & 0.1 \\
Regularizer $\lambda$ & 0.1  & 0.4 & 0.6  \\
Hidden neurons & 12 & 12  & 12 \\
$N_{conn.,1}$ & 12  & 10  & 8 \\
$N_{conn.,2}$ & 6 & 6 & 4 \\
Patience (Genetic algorithm) & 5 & 5 & 5 \\
Tolerance (Genetic Algorithm) & $1\mathrm{e}{-4}$  & $1\mathrm{e}{-4}$  & $1\mathrm{e}{-4}$  \\
Learning rate & 0.03  & 0.1 & 0.1 \\ 
Patience (Early Stopping) & 5  & 25  & 20  \\
Tolerance (Early Stopping) & $1\mathrm{e}{-4}$  & $1\mathrm{e}{-6}$  & $1\mathrm{e}{-6}$  \\\hline
\end{tabular}%
}
\end{table}
There are two different parameters called \textit{patience} and \textit{tolerance} that are used as stopping criteria if a value does not change within some tolerance for a certain time (patience). One refers to the number of epochs and the threshold used for Early Stopping, while the second is used as termination criterion for the genetic algorithm.
The code is available at \url{https://github.com/jspieler/QBAF-Learning}.

\subsection{Results}

\begin{figure}
\centering
\includegraphics[width=\textwidth]{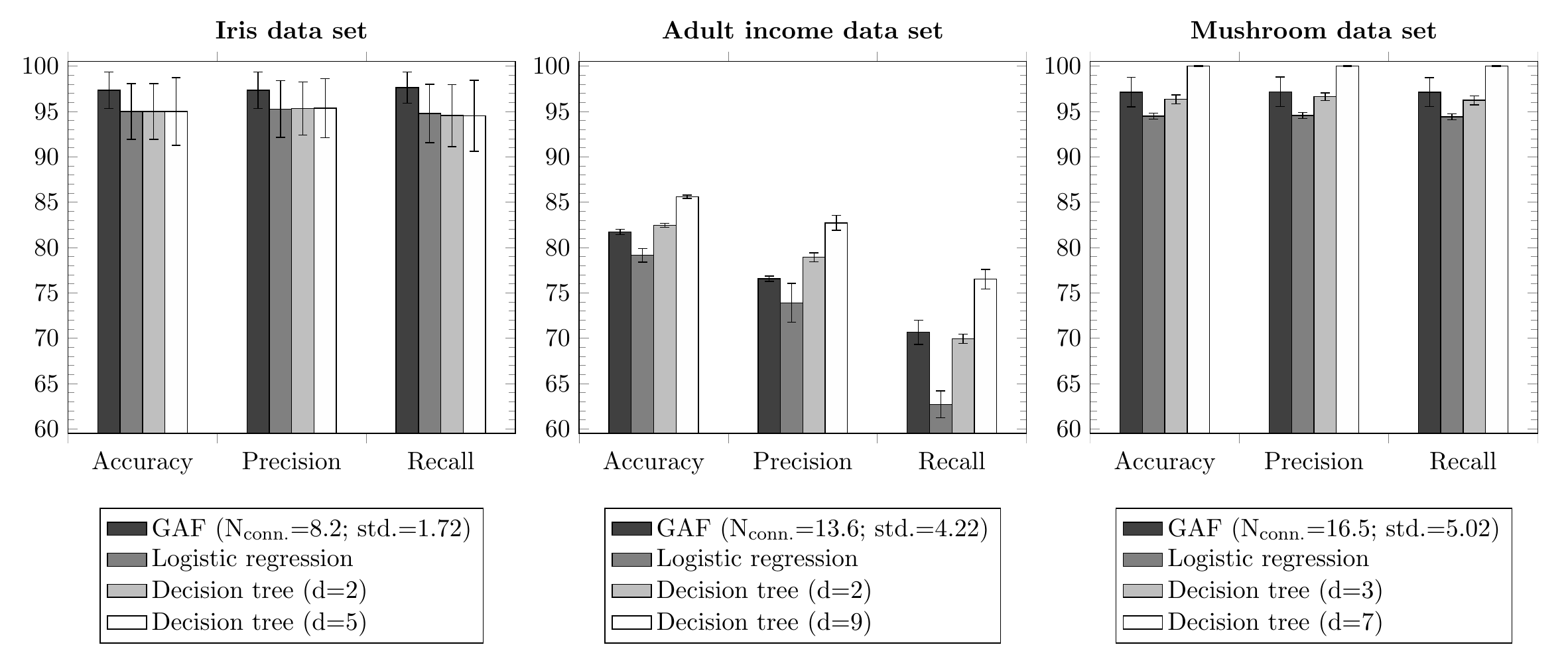}
\caption{Bar graphs representing the means for accuracy, precision and recall $(\%)$ of GAFs, logistic regression and decision trees for different data sets (test data). Error bars indicate the corresponding standard deviation.}
\label{fig:results_bar_chart}
\end{figure}

Figure \ref{fig:results_bar_chart} illustrates our experimental results in a
bar chart.
We report each of the metrics with mean and standard deviation for $10$ runs since there is randomness during training that influences the results. Each data set is split into training, validation and test data by a ratio of $70/10/20$.

\begin{figure}[tb]
\centering
\includegraphics[width=\textwidth]{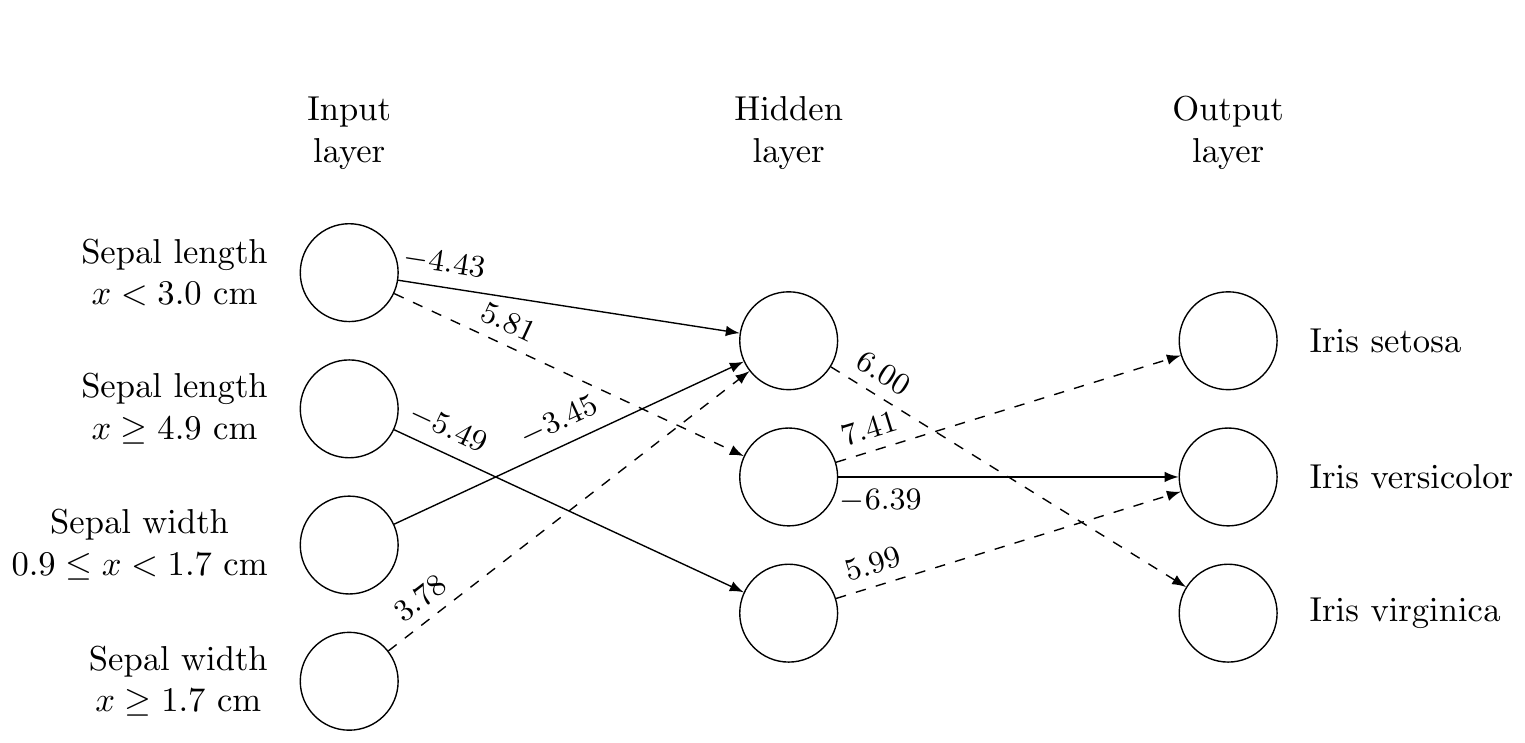}
\caption{Example of a classification GAF for the iris data set.}
\label{fig:AF_iris}
\end{figure}

For the iris data set, the argumentation classifiers achieve an test accuracy that is about $2~\%$ higher than those achieved by logistic regression and decision trees. The number of connections has a mean of $8.2$ which leads to well-interpretable classifiers. An example of an classification GAF for the iris data set is shown in figure \ref{fig:AF_iris}.
As before, attack relations are represented by solid and support relations by dashed edges. The weights indicate how strongly an argument attacks or supports another one. For example, a sepal length smaller than $3.0$ centimetres supports the iris species ``setosa'' and attacks the species ``versicolor''. 

For the adult income data set, the results of GAFs are about $2~\%$ better than those of logistic regression. However, decision trees achieve a higher accuracy than GAFs. 

For the mushroom data set, GAFs perform better than logistic regression and
similar to a decision tree of depth $3$. Increasing the depth of a decision tree leads to an accuracy of $100~\%$. 

\begin{figure}
\centering
\includegraphics[width=\textwidth]{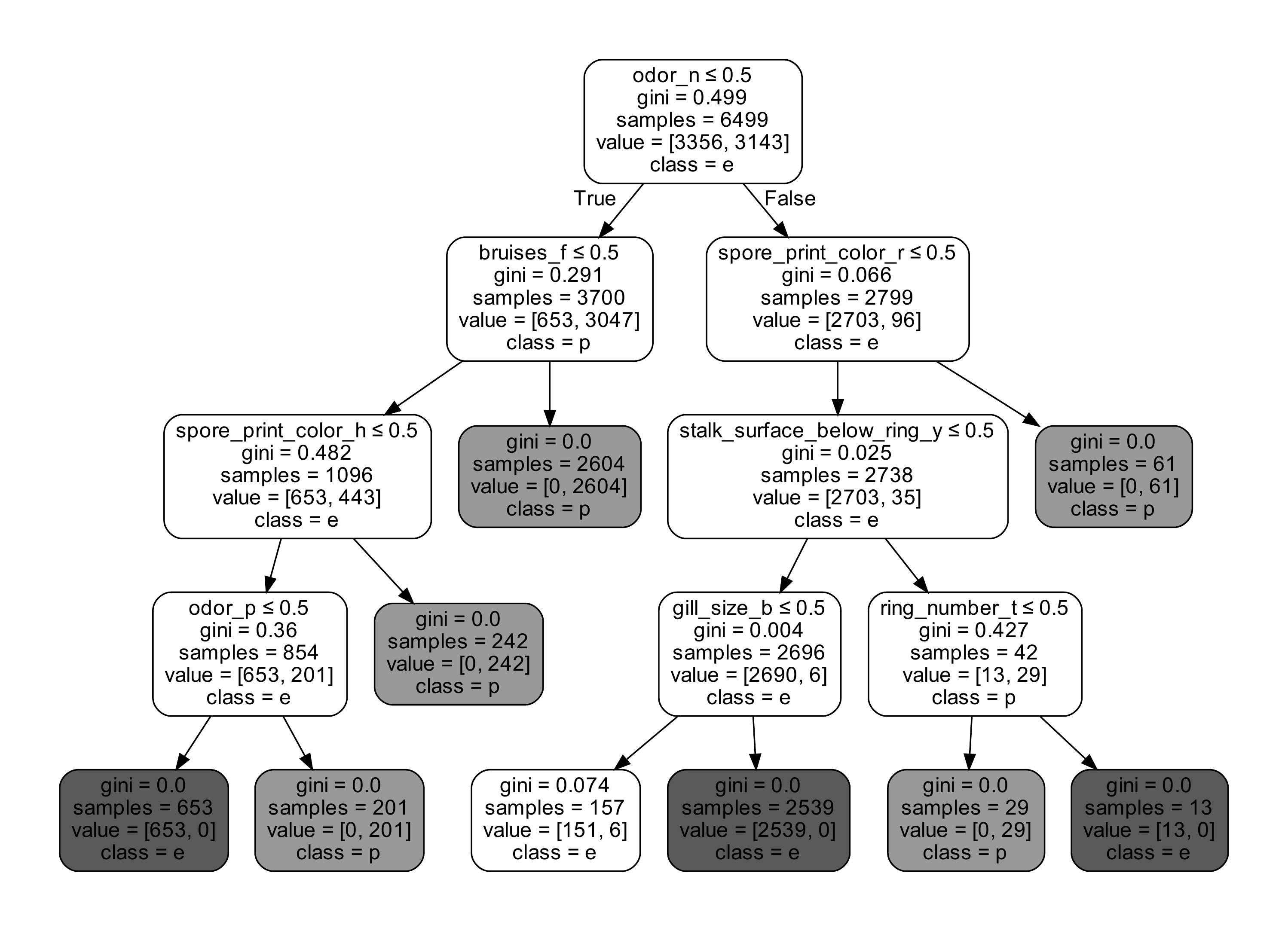}
\caption{Example of a decision tree classifier for the mushroom data set.}
\label{fig:DecisionTreeMushroom}
\end{figure}

\begin{figure}[tb]
\centering
\includegraphics[width=\textwidth]{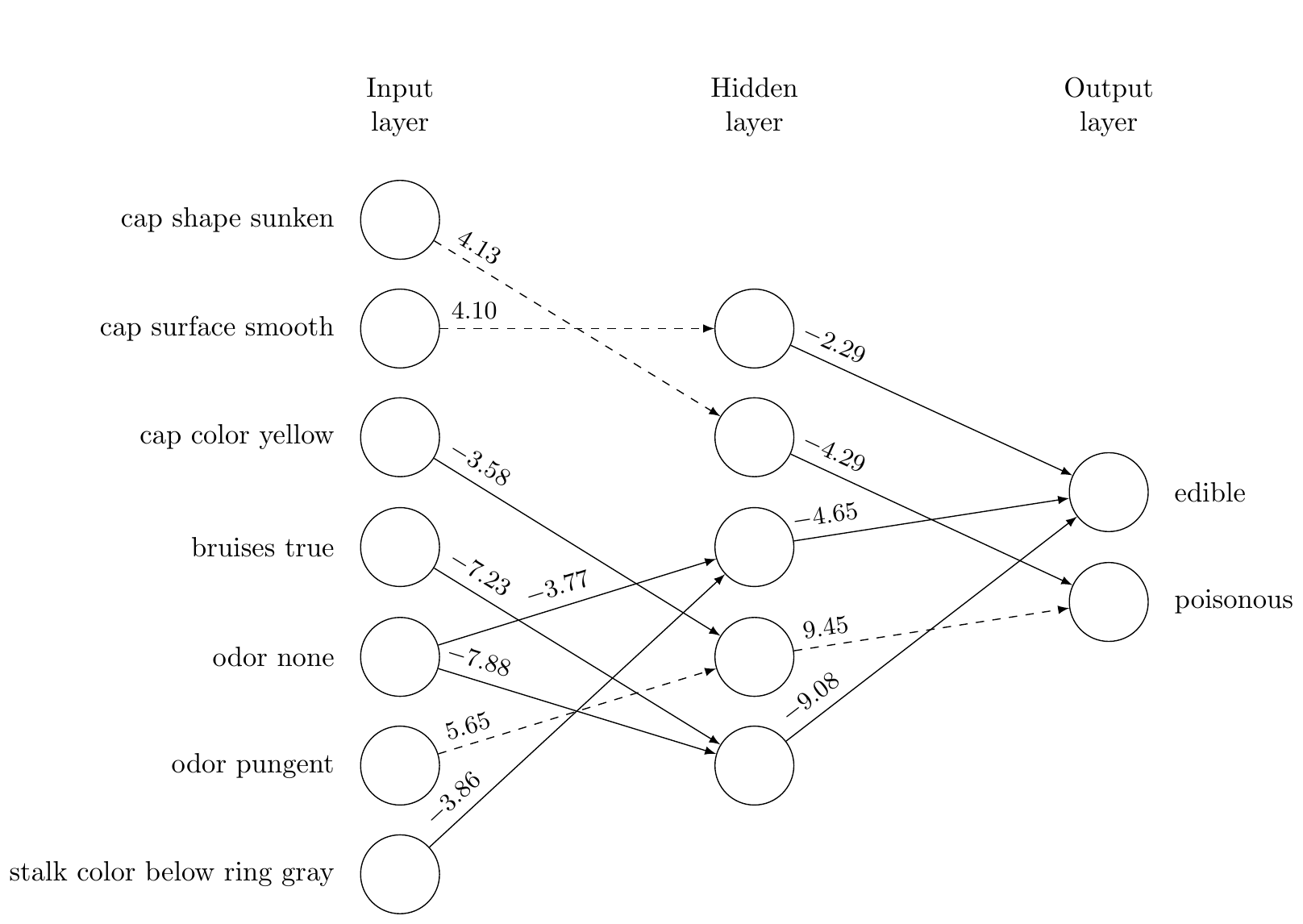}
\caption{Example of a classification GAF for the mushroom data set.}
\label{fig:AF_mushroom}
\end{figure}

On average, GAFs perform similar to flat decision trees.  
Deep decision trees achieve a significantly better performance on the adult 
income and the mushroom data set. However, deep decision trees are not well interpretable anymore. Since even a decision tree with depth $7$ is very large,
we show only a decision tree of depth $4$ for the mushroom data set in Figure \ref{fig:DecisionTreeMushroom}.
Even for this depth, tracing classes back to corresponding inputs is difficult because there are already $5$ leaf nodes for the class ``poisonous'' (p) and another $4$ for ``edible'' (e).
For direct comparison, Figure \ref{fig:AF_mushroom} shows a GAF for the 
mushroom dataset.

\section{Related Work}

In recent years, there has been increasing interest in the combination of argumentation frameworks and machine learning methods \cite{Cocarascu_2018}. Most of the approaches first apply machine learning methods to extract arguments and their relations and afterwards solve the argumentation problem \cite{cocarascu2020data,Cocarascu_2016,Lippi_2015,thimm2017towards}. Although this has been successfully used for different applications, e.g. online reviews \cite{Cocarascu_2016,Cocarascu_2018}, recommendations systems \cite{Rago_2018} or fake news detection \cite{Cocarascu_2018}, one potential problem is that
the machine learning and argumentation steps are considered independently.
In contrast, we learn argumentative classifiers in an end-to-end fashion
for the classification task at hand. 

There has also been some work on using neural networks for argumentation. Kuhlmann and Thimm \cite{Kuhlmann_2019} proposed to use \textit{graph convolutional networks} to approximate the acceptance of arguments under different semantics. The decision of acceptability is modeled as a classification problem. Craandijk and Bex \cite{Craandijk_2020} proposed to use so-called \textit{argumentation graph neural networks} (AGNN), to predict the acceptability of arguments under different semantics. A function that maps the input to a binary label (accepted/rejected) is approximated by computing the likelihood for an argument to be accepted and round it to a value of either 0 or 1. The model learns a message passing algorithm to predict the likelihood that an argument is accepted. \cite{RiveretTG20} recently trained Boltzman machines
on argument labellings such that the model can generate and predict
correct labellings.

Since GAFs can be seen as sparse MLPs, there are also relationships to
learning sparse neural networks. However, sparse neural networks are usually
not motivated by the desire to learn an interpretable network, but
to decrease the risk for overfitting, the memory and runtime complexity of
learning neural networks and the associated power consumption.
Although deep neural networks have proven to be extremely successful in many applications, the aformentioned reasons lead to growing interest in sparse neural networks. There are a lot of approaches, including pruning of weights and/or connections \cite{LeCun_1990,Han_2016,Han_2015}, compression \cite{Chen_2015,Guo_2016,Hinton_2015,Louizos_2017}, quantization \cite{Courbariaux_2015,Courbariaux_2016,Han_2016,Lin_2016,Zhou_2016} and low rank approximation \cite{Denil_2014,Jaderberg_2014,Liu_2015,Tai_2016}. 
However, most of the approaches rather focus on reducing overfitting, memory and computational demand than on getting sparsely-connected and interpretable networks. Furthermore, techniques such as pruning or quantization require the whole network to be trained before removing parameters \cite{Dey_2018}. Usually, sparsely connected layers are implemented by fully-connected layers whose sparsity is enforced via binary mask matrices \cite{Ardakani_2017,Liu_2020}. \cite{Liu_2020} also proposed a method to use sparse data structures. 

Historically, genetic algorithms were also used as an alternative to gradient descent-based parameter learning for neural networks architectures \cite{Honavar_1995,De_Jong_1988,Miller_1991,Montana_1989}. 
Designing neural network architectures is often complex because there are a lot of degrees of freedom. 
Often, individual experience, knowledge about the application domain or 
heuristics are used to define the structure of neural networks. 
Miller, Todd and Hegde \cite{Miller_1991} proposed to use genetic algorithms to automatically design neural networks. In recent years, a research field called \textit{Neuroevolution} has developed that studies the use of evolutionary algorithms to create neural networks. The learning process either includes evolving the weights, the structure or both. Sometimes the hyperparameters are also part of the learning process. A popular algorithm is \textit{NeuroEvolution of Augmenting Topologies} (NEAT) \cite{Stanley_2002} that uses genetic algorithms to evolve the parameters and the architecture of neural networks.
Each network starts with no hidden neuron an is evolved by gradually adding new nodes and connections. Many variations of NEAT have been developed over the years. 
There are also some recent approaches that combine Neuroevolution for structure learning with Backpropagation and gradient descent for weight optimization of neural networks \cite{Elsken_2019,Stanley_2019}.

\section{Conclusions and Future Work}

We proposed a first algorithm for learning classification GAFs in an
end-to-end fashion from data. Our experiments show that they already perform
similar to decision trees in terms of learning performance and interpretability.
An advantage over decision trees is perhaps that the classification
decision occurs only once in the output layer. In contrast, in a decision tree,
the same class label can often be reached by multiple paths. This can make it more difficult to recognize relationships between features.

There are several ways to potentially increase the performance.
On the algorithmic side, we can replace the general-purpose operators in
our genetic algorithm with operators that are tailored to the special
structure that we want to learn. We are also investigating alternative
meta-heuristics like swarm algorithms.
Furthermore, often simplification rules can be applied to the final GAF. 
For example, attackers which are in turn attacked by another argument can
potentially be simplified by a support relation between the corresponding arguments. This can increase the interpretability of the learned GAFs further. 

On the representation side, it may be beneficial to add more argumentative
structure both for learning performance and interpretability.
For example, the idea of collective attacks 
\cite{bodanza2017collective,nielsen2006generalization} that 
capture the joint effect of multiple arguments in a single edge
can be helpful to capture more complex relationships in a
sparser graphical strucuture. However, the semantics of these
collective attacks should ideally be defined by a differentiable function
in order to keep the learning problem simple. 
Another potentially interesting idea is to consider fuzzy arguments
rather than binarized arguments. For example, in case of a numerical attribute,
instead of introducing input arguments that accepted for particular value ranges, we may consider a fuzzy argument that fully accepts at the center of
the interval and gradually decreases the degree of acceptance as the value 
approaches the boundaries of the interval.




\bibliographystyle{splncs04} 
\bibliography{references}

\end{document}